\documentclass{article}

\usepackage{arxiv}

\usepackage[utf8]{inputenc} 
\usepackage[T1]{fontenc}    
\usepackage{hyperref}       
\usepackage{url}            
\usepackage{booktabs}       
\usepackage{amsfonts}       
\usepackage{nicefrac}       
\usepackage{microtype}      
\usepackage{lipsum}
\usepackage{graphicx}
\graphicspath{ {./images/} }
\usepackage{float}
\usepackage{multirow}

\title{MetaGen Blended RAG: Unlocking Zero-Shot Precision for Specialized Domain Question-Answering}

\author{
  Kunal Sawarkar \\
   IBM \\
  \texttt{Kunal@ibm.com} \\
    \And
  Shivam R Solanki  \\
  IBM \\
  \texttt{Shivam.Raj.Solanki@ibm.com} \\
  \And
  Abhilasha Mangal \\
  IBM \\
  \texttt{Abhilasha.Mangal@ibm.com} \\
}

\begin{document}

\maketitle

\begin{abstract}
Retrieval-Augmented Generation (RAG) struggles with domain-specific enterprise datasets, often isolated behind firewalls and rich in complex, specialized terminology unseen by LLMs during pre-training. Semantic variability across domains like medicine, networking, or law hampers RAG's context precision, while fine-tuning solutions are costly, slow, and lack generalization as new data emerges. Achieving zero-shot precision with retrievers without fine tuning still remains a key challenge. We introduce 'MetaGen Blended RAG', a novel enterprise search approach that enhances semantic retrievers through a metadata generation pipeline and hybrid query indexes using dense and sparse vectors. By leveraging key concepts, topics, and acronyms, our method creates metadata-enriched semantic indexes and boosted hybrid queries, delivering robust, scalable performance without fine-tuning. On biomedical PubMedQA dataset, MetaGen Blended RAG achieves 82\% retrieval accuracy and 77\% RAG accuracy, surpassing all prior zero-shot RAG benchmarks and even rivaling fine-tuned models on that dataset, while also excelling on datasets like SQuAD and NQ. This approach redefines enterprise search using new approach to building semantic retrievers with unmatched generalization across specialized domains.
\end{abstract}

\section{Introduction}

Domain-specific question answering (QA) systems have become a cornerstone of applied AI. Retrieval-Augmented Generation (RAG) represents a paradigm shift in natural language processing, integrating the strengths of information retrieval and generative models to address the limitations of LLMs. Unlike standalone LLMs, which rely solely on parametric knowledge encoded during pre-training, RAG enhances contextual understanding by retrieving relevant external documents to inform response generation \cite{lewis2020retrieval}. This approach minimizes issues such as hallucination and outdated information, making RAG particularly effective for knowledge-intensive tasks.

The efficacy of RAG primarily hinges on the performance of its retrieval component, which must efficiently identify and rank documents that are semantically and contextually aligned with the user query. However, challenges arise in domain-specific settings where corpora often lack metadata or contain complex terminology unseen during LLM training \cite{gao2023retrieval}. Traditional retrieval methods struggle with semantic variability and metadata-deficient environments, leading to suboptimal context precision. To address these issues, advanced indexing and retrieval strategies have been developed, ranging from lexical to semantic and hybrid approaches, each offering unique trade-offs in accuracy, scalability, and computational efficiency. 

\paragraph{Search Indexing Techniques for Retrievers}

Enterprise Search indexing forms the backbone of the retrieval component in RAG systems, determining how effectively relevant documents are identified. Lexical indexing, such as the BM25 algorithm, relies on term frequency and inverse document frequency to rank documents based on keyword matches \cite{robertson2009probabilistic}. BM25 excels in scenarios with structured queries and high term overlap. However, its reliance on exact term matching limits its ability to capture semantic relationships, making it less effective for queries with paraphrased or conceptually related terms, a common challenge in domain-specific corpora.

Semantic indexing, in contrast, leverages dense or sparse vector representations generated by transformer-based models, such as BERT or sentence transformers, to encode documents and queries into a shared embedding space \cite{reimers2019sentence}. Certain semantic indexes employ techniques like k-nearest neighbors (k-NN) search \cite{malkov2018efficient},  capturing deeper contextual and semantic similarities, enabling robust performance in scenarios with low term overlap \cite{karpukhin-etal-2020-dense}. However, it is computationally intensive and underperforms in metadata-deficient settings where embeddings lack sufficient discriminative power. Hybrid indexing combines lexical and semantic approaches, using score fusion techniques (e.g., weighted averaging of BM25 and cosine similarity scores) to balance precision and recall \cite{lin2021dense}. Hybrid methods have shown superior performance in diverse datasets, as they mitigate the weaknesses of individual approaches while enhancing the accuracy of the retrieval of RAG systems \cite{sawarkar2024blendedrag}, but do not yet provide enough efficacy for highly metadata-deficient corpora.

\section{Related Work}

\paragraph{Challenges with a domain-specific QA}

PubMedQA is considered one of the most challenging dataset for the domain specific problems. Model has to answer biomedical research questions by reasoning over complex abstracts. It requires understanding specialized terminology and interpreting evidence from the abstract’s text to produce yes/no/maybe answers. With only a small set of expert-labeled examples amid many unlabeled ones, the dataset tests a model’s ability to handle limited data while performing deep, domain-specific reasoning. The test set has 1000 Q\&A pairs and 62000 documents. 

The domain specific corpora, such as PubMedQA, contain multiple passages with information about the same topic (disease/condition). So, the retrieval based on keywords and semantic search can struggle to identify the right document as well as the right passage within the document. Additionally, the LLM is unable to use the right passage from the list of passages talking about the same disease/condition to generate/conclude the final answer. In summary it suffers from challenges of Multi-dimensionality, data sparsity, and multiple interpretations of terms based on the context. 

The challenge of building high-accuracy RAG pipelines in domain-specific settings, such as biomedical, has led to the emergence of two dominant strategies: fine-tuning-based approaches and retriever-focused methodologies. 

\paragraph{Fine-Tuning-Based RAG Systems.}
Several recent efforts have attempted to fine-tune LLMs for RAG pipelines specifically for biomedical or clinical QA tasks. Notable among these are AlzheimerRAG \cite{lahiri2024alzheimerrag}, which integrates multimodal data and achieves 74.0\% accuracy on PubMedQA \cite{lahiri2024alzheimerrag}, and RAFT \cite{zhang2024raft}, which applies retrieval-augmented fine-tuning (RAFT) using LLaMA2-7B and reaches 73.3\%. Similarly, DSF + RAG employs domain-specific fine-tuning techniques to score 71.6\%, while MEDRAG + GPT-4 leverages optimized medical retrieval to attain 70.6\%. These approaches demonstrate that fine-tuning can effectively adapt a model's reasoning capabilities to domain-specific linguistic patterns and factual grounding requirements.

However, the success of such fine-tuned systems is tightly coupled to the quality of the tuning set and the availability of in-domain labeled data. In other enterprise use cases for metadata-sparse environments—where structured descriptors like MeSH terms or section titles are limited—fine-tuning alone cannot compensate for poor retrieval quality. Moreover, these methods often require considerable computational \& human resources and may struggle to generalize outside the specific distributions they were tuned on. This limits their scalability and maintainability, especially in real-world enterprise applications where corpus characteristics can evolve rapidly.

\paragraph{Retriever-Based RAG Systems.}
In contrast, retrieval-centric RAG systems attempt to improve accuracy by enhancing the retriever's ability to fetch semantically relevant context passages without altering the language model. For instance, GPT-3.5 + RAG, which combines OpenAI's GPT-3.5 with a standard retriever, achieves a respectable 71.6\% accuracy. LLaMA2-7B + RAG, a zero-shot RAG pipeline using LLaMA2-7B without any domain adaptation, trails significantly at 58.8\% accuracy. These results highlight a critical bottleneck: without domain-specific fine-tuning or enhanced indexing, even strong language models struggle to generate accurate answers when fed with suboptimal context retrieved from unstructured or semantically ambiguous corpora.

Retriever-based systems are attractive due to their modularity, lower computational cost, and ease of deployment. Yet, their effectiveness is severely limited in settings where the retrieval component cannot semantically bridge domain-specific terms, acronyms, or variations in medical language. In such cases, traditional lexical search or dense retrieval falls short, as these models lack the contextual understanding needed to surface the most relevant content.

\section{Methodology}
\paragraph{Motivation for Metadata Enrichment.}
In Domain Specific QA, questions are highly specialized, and correct answers are often buried within long, unstructured abstracts. In such contexts, achieving high RAG accuracy depends on both the granularity of the retriever and its ability to leverage domain-specific linguistic features. As a result, improving document representations through metadata augmentation has emerged as a promising direction, especially in settings where fine-tuning is infeasible or metadata is limited.

We have chosen this direction, as this does not hinge on the need to fine-tune the model with expensive annotation sets. Furthermore, the approach is complementary to any model that is already fine-tuned and can further enhance the search accuracy. Most real-world document corpora do not have any organic metadata tagged to them. Such metadata metadata-deficient corpus does not benefit from indexing optimization. Generation of metadata can also be be very noisy and thereby drown the relevant parts with irrelevant keywords.  

We therefore focused our efforts on the design space of domain-adaptive RAG systems using a hybrid retriever with blended queries and metadata-aware indexing, which pushes the boundary of performance in metadata-deficient, domain-intensive environments. Our investigation focuses on evaluating both retrieval and generative accuracy in the context of the PubMedQA dataset—a benchmark that demands precise retrieval of specialized medical knowledge and factually grounded answer generation. As we demonstrate in subsequent sections, metadata-enriched retrieval pipelines yield state-of-the-art RAG performance while preserving generalizability, scalability, and efficiency.

\subsection{MetaGen Infused Retrieval-Augmented Generation Pipeline}

\begin{figure}[h]
\centering
\includegraphics[width=1\linewidth]{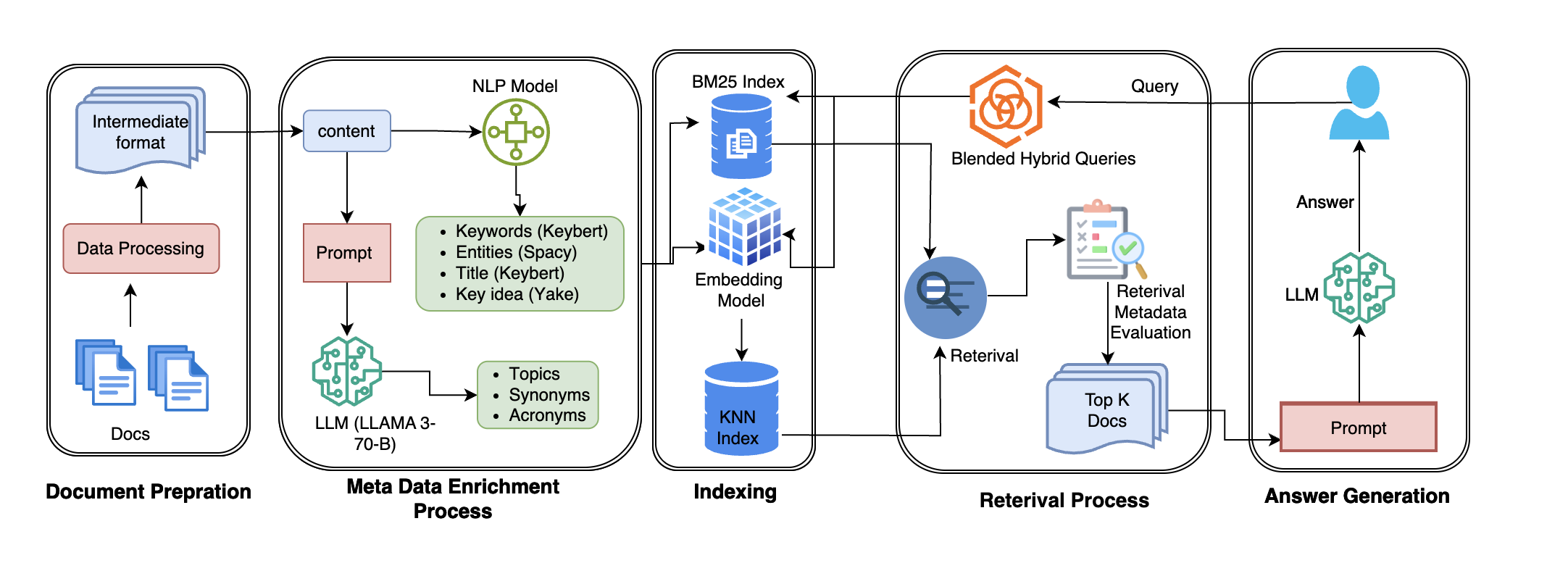}
\caption{MetaGen Blended Retrieval-Augmented Generation Pipeline}
\label{fig:MGRP}
\end{figure}

The proposed MetaGen Blended Retrieval-Augmented Generation(MBRAG) pipeline comprises a systematic five-stage workflow that integrates metadata-driven enrichment and hybrid retrieval techniques as presented in Figure \ref{fig:MGRP}.

The pipeline initiates with the \textbf{ingestion stage}, where raw documents are batch-loaded or dynamically acquired from relevant sources.

Following ingestion, the pipeline executes a critical \textbf{metadata enrichment stage}. In this phase, metadata streams are conditionally extracted and systematically calibrated based on domain-specific criteria and optimized according to performance metrics described in Section 3.2. This selective approach to metadata enrichment ensures computational efficiency and relevance, as only metadata elements that demonstrably contribute to improved retrieval precision are retained.

The metadata-enriched corpus is subsequently indexed during the \textbf{indexing stage}, employing a dual indexing strategy designed to capture both lexical and semantic dimensions of the documents. Specifically, a traditional BM25-based index is established to facilitate rapid and accurate keyword retrieval across raw textual data and associated metadata fields. Concurrently, a dense vector index is constructed by computing embeddings that encapsulate deeper semantic relationships inherent within the enriched document representations.

During the \textbf{retrieval stage}, the pipeline deploys a hybrid retrieval strategy that simultaneously leverages both lexical (BM25-based) and semantic (vector-based k-NN) retrieval mechanisms. The outputs from these parallel searches are combined through score normalization techniques, employing a weighted fusion of BM25 and cosine similarity scores to yield a refined, contextually pertinent document ranking. This hybrid approach significantly enhances the relevance and precision of retrieved documents, particularly within domain-specific corpora.

Finally, the pipeline culminates in the \textbf{answer generation stage}, where top-ranked documents from the retrieval stage are aggregated and presented to an LLM in the form of structured prompts. The model synthesizes contextual information from these high-quality retrievals to generate precise and coherent answers.

By integrating conditional metadata enrichment, strategic indexing methodologies, and hybrid retrieval mechanisms, the proposed pipeline effectively scales to extensive, specialized datasets. 

\subsection{Metadata Enrichment Details}

\begin{figure}[h]
\centering
\includegraphics[width=1\linewidth]{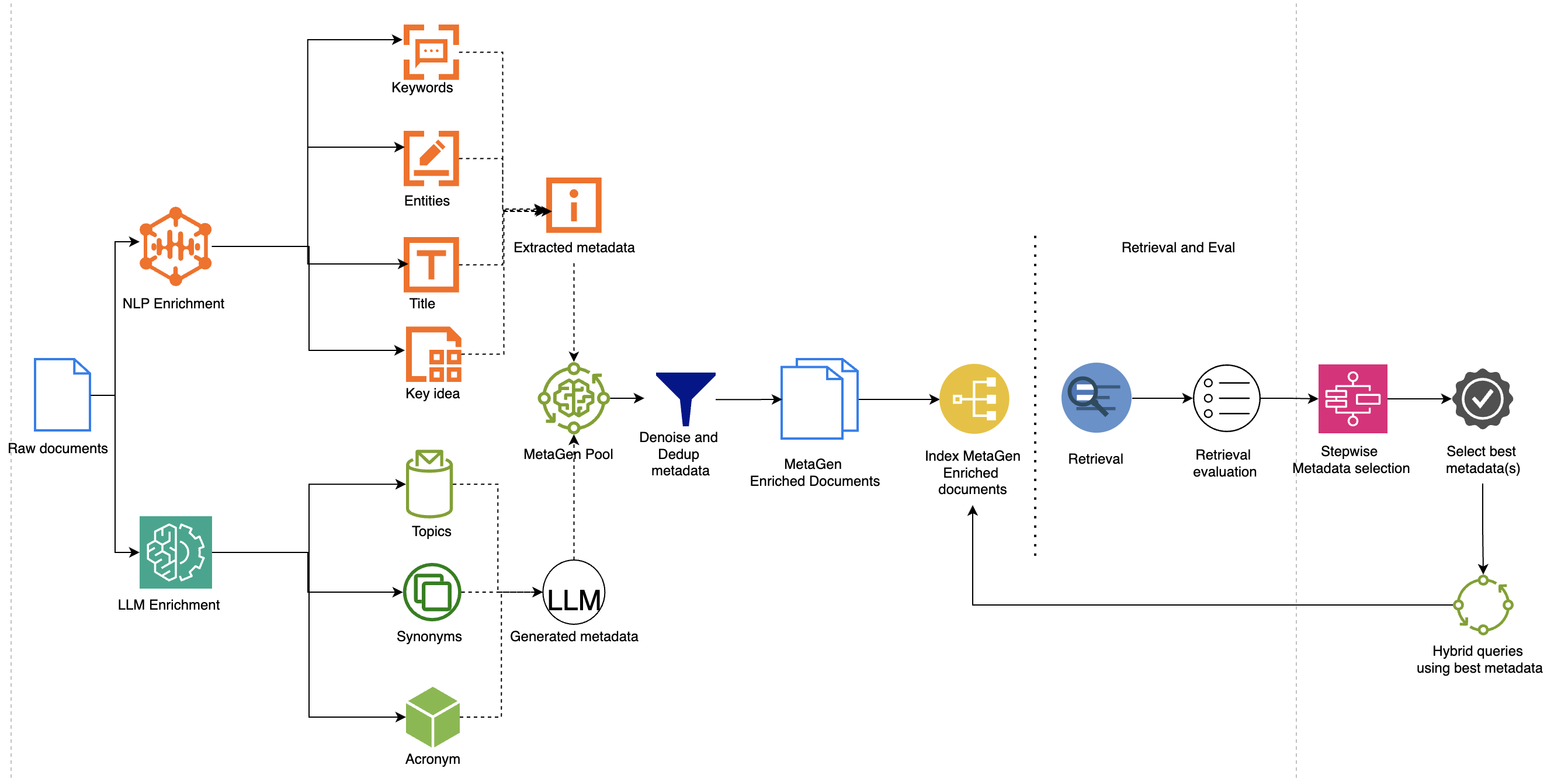}
\caption{MetaGen Enrichment Workflow}
\label{fig:MEGF}
\end{figure}

Our proposed metadata enrichment framework comprises two integrated phases: a \textit{conditional metadata enrichment} phase followed by a systematic \textit{stepwise metadata selection} process, as illustrated in Figure \ref{fig:MEGF}. These two stages collectively ensure a computationally efficient yet semantically robust approach to metadata enrichment, optimized specifically for RAG.

\subsubsection{Conditional Metadata Enrichment}

The conditional metadata enrichment stage balances semantic depth with computational efficiency. It selectively executes computationally intensive enrichment processes based on corpus size thresholds. Initially, raw documents from various formats (e.g., PDF, HTML) are uniformly transformed into structured plaintext. Subsequently, a conditional logic assesses corpus scale: if the total corpus size, denoted as $|D|$, exceeds a predefined threshold (typically in the millions), the enrichment pipeline omits the computationally expensive LLM-driven metadata extraction step. Conversely, smaller corpora undergo full enrichment across all metadata extraction streams.

Independent of corpus size, certain NLP-driven enrichment steps are always executed to ensure baseline metadata coverage:

\textbf{KeyBERT Extraction \cite{issa2023comparative} }Leveraging a sentence-transformer backbone, this step identifies a single high-impact keyphrase and title from each document, capturing essential semantic nuances efficiently.

\textbf{YAKE Extraction \cite{gupta2024natural}:} Utilizing statistical heuristics based solely on local textual properties, this step generates one representative phrase encapsulating the central idea of each document.

\textbf{spaCy Named Entity Recognition\cite{kumar2023algorithm}:} This component extracts the three most prominent named entities from each document across predefined entity classes (e.g., PERSON, ORG, LOCATION).

For corpora below the predefined threshold, an additional high-precision enrichment step employing a large instruction-tuned LLM (e.g., meta-llama/llama-3-70B-instruct) is activated. This LLM-based enrichment phase entails prompting the model to extract comprehensive, domain-specific tags including topical keywords, significant phrases, synonyms, and acronyms, structured in JSON format for seamless integration. An illustrative prompt structure is as follows:

\begin{quote}
\textbf{System Prompt:} \textit{"You are a helpful, respectful, and truthful assistant. Avoid generating or inferring unsupported information, maintaining ethical and factual accuracy."}

\textbf{User Prompt:} \textit{"You are a medical domain expert. Extract key topics, important phrases, synonyms, and acronyms from the provided medical text: \{context\}. Return outputs strictly in valid JSON format with the keys: topics, phrases, synonyms, and acronyms, excluding any extraneous information."}

\end{quote}

Outputs from all enrichment streams (NLP-based and, when applicable, LLM-based) are consolidated into a unified, deduplicated repository termed the \textit{MetaGen Pool}. This repository undergoes further denoising procedures to ensure metadata integrity, accuracy, and relevance for downstream retrieval processes.

\subsubsection{Stepwise Metadata Selection}

The subsequent stepwise metadata selection phase systematically determines the important subset of metadata streams from the enriched \textit{MetaGen Pool}. This iterative, data-driven methodology utilizes a forward selection approach to empirically identify metadata streams that yield significant retrieval accuracy improvements, as depicted in Figure \ref{fig:MEGF}.

The process initiates with an empty metadata set evaluated against a predefined ground-truth retrieval benchmark. In each iteration, candidate metadata streams (KeyBERT, YAKE, NER, LLM-derived tags) are individually considered. The specific candidate stream under evaluation is provisionally integrated into hybrid queries used for retrieval, and the resulting improvement in retrieval accuracy (measured by recall@k) is rigorously evaluated against the baseline.

Streams contributing statistically significant retrieval gains are retained, while those failing to yield measurable improvements are systematically discarded. This iterative evaluation continues until no additional metadata stream provides substantial enhancement. The final set of selected metadata streams constitutes the optimal subset leveraged by the RAG pipeline.

This adaptive and rigorous metadata selection strategy ensures maximal retrieval performance benefits while concurrently preventing unnecessary computational expenditures, thus ensuring both scalability and efficiency within large-scale, enterprise-grade retrieval systems.

\section{Experiments \& Results}

\subsection{Retriever Evaluation}

To assess the efficacy of our method, we conducted extensive experiments using the domain-specific PubMedQA dataset. We constructed a corpus combining approximately 62,000 documents derived from PubMedQA, which were subsequently enriched through our proposed metadata augmentation pipeline. This systematic metadata enrichment was specifically designed to enhance document retrieval accuracy, a critical component influencing overall RAG performance. The retrieval effectiveness was evaluated by measuring the accuracy of retrieving the correct document at the top position ($k=1$) across various retrieval strategies.

Figure \ref{fig:MBRA} illustrates the progressive improvement in retrieval accuracy across different retrieval configurations:

\begin{figure}[h]
\centering
\includegraphics[width=0.8\linewidth]{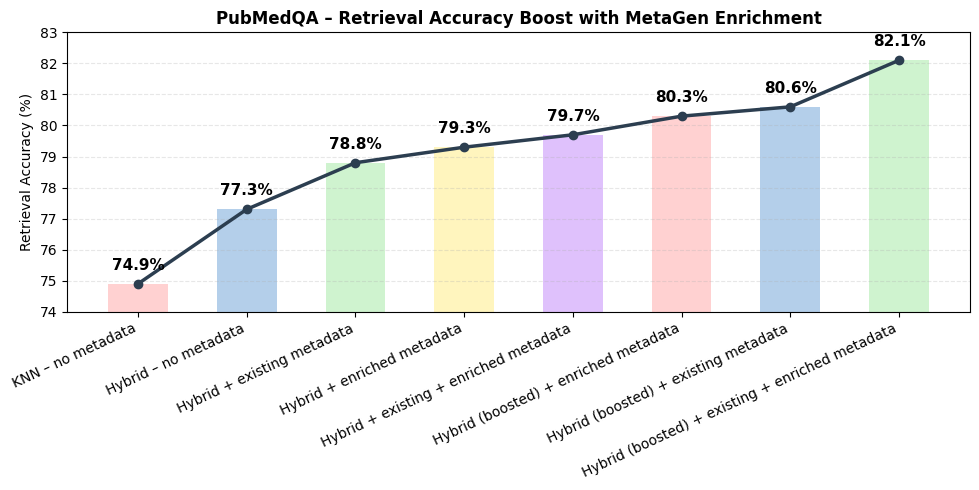}
\caption{Impact of Metadata Enrichment on Retrieval Accuracy (PubMedQA dataset)}
\label{fig:MBRA}
\end{figure}
Initially, we established a baseline using a plain k-nearest neighbors (k-NN) semantic search without incorporating any metadata. This baseline achieved a retrieval accuracy of approximately 74.9\%. Subsequently, integrating lexical (BM25) search with semantic (k-NN) search into a hybrid retrieval framework improved the accuracy to 77.3\%, demonstrating the complementary advantages of combining lexical and semantic retrieval methods even in the absence of additional metadata.

To quantify the impact of metadata, we progressively augmented the retrieval framework with both existing metadata and our enriched metadata streams. Incorporating existing metadata fields—such as Medical Subject Headings (MeSH terms) —yielded a notable accuracy improvement, elevating performance to 78.8\%. When further supplemented by enriched metadata fields derived through NLP techniques (e.g., keyphrases, named entities) and LLM-generated semantic annotations (topics, synonyms, acronyms), retrieval accuracy improved incrementally, reaching 79.3\% with enriched metadata alone and 79.7\% when combined with existing metadata.

To maximize retrieval performance, we implemented field-boosted configurations, selectively emphasizing critical metadata fields during the query formulation. Using boosted queries with only enriched metadata improved accuracy to 80.3\%, whereas employing boosted queries with only existing metadata achieved 80.6\%. Notably, the highest accuracy of 82.1\% was attained by applying boosted hybrid retrieval that leveraged both existing and enriched metadata simultaneously. This final configuration represents a substantial absolute gain of approximately 7.2 percentage points over the hybrid search baseline without metadata, and approximately 10 points above the initial semantic (k-NN) baseline.

These empirical results clearly indicate that targeted metadata enrichment combined with hybrid query boosting substantially enhances retrieval precision. Such significant retrieval gains consequently facilitate downstream improvements in RAG performance.

This underscores the viability and efficacy of our approach for improving document retrieval in specialized, metadata-deficient corpora

\subsubsection{Evaluation on Natural Questions (NQ) Dataset}

To validate the generalizability of our metadata enrichment approach beyond domain-specific settings, we conducted experiments on the widely used Natural Questions (NQ) dataset. Specifically, we constructed a large-scale corpus comprising approximately 5 million documents derived from NQ and systematically enriched this corpus with our proposed MetaGen metadata pipeline. Retrieval performance was assessed using the standard top-5 accuracy metric, allowing us to measure the practical benefit of metadata augmentation in a generic retrieval scenario.

As illustrated in Table \ref{tab:retrieval-improvements} the MetaGen enrichment significantly enhanced retrieval performance. Incorporating enriched metadata into the retrieval process resulted in an absolute improvement of 20.98 percentage points in top-5 retrieval accuracy compared to the baseline configuration without metadata. This marked performance gain highlights the broad applicability of the approach, even when applied to generic and diverse question-answering datasets such as NQ.

\begin{table}[h]
\caption{Retrieval accuracy improvements using MetaGen RAG across datasets}
\label{tab:retrieval-improvements}
\centering
\begin{tabular}{@{}p{1.8cm}p{7.2cm}c@{}}
\toprule
\textbf{Dataset} & \textbf{Search + Metadata Variant} & \textbf{Retrieval Accuracy (\%)} \\
\midrule
\multirow{8}{*}{\textbf{PubMedQA}} 
    & KNN search without any metadata & 74.9 \\
    & Hybrid search without any metadata & 77.3 \\
    & Hybrid search only with existing metadata & 78.8 \\
    & Hybrid search only with enriched metadata & 79.3 \\
    & Hybrid search with existing + enriched metadata & 79.7 \\
    & Hybrid (boosted) with enriched metadata & 80.3 \\
    & Hybrid (boosted) with existing metadata & 80.6 \\
    & Hybrid (boosted) with existing + enriched metadata & \textbf{82.1} \\
\midrule
\multirow{3}{*}{\textbf{NQ}} 
    & Without metadata & 49.99 \\
    & With existing metadata & 59.49 \\
    & Existing + enriched metadata & \textbf{60.48} \\
\midrule
\multirow{3}{*}{\textbf{SQuAD}} 
    & Without metadata & 93.30 \\
    & With existing metadata & 93.58 \\
    & Existing + enriched metadata & \textbf{93.68} \\
\bottomrule
\end{tabular}
\end{table}

\subsubsection{Evaluation on SQuAD Dataset}

We further extended our evaluation to the Stanford Question Answering Dataset 
(SQuAD)~\cite{rajpurkar2016squad}, a widely-adopted benchmark for question-answering systems. Due to its relatively structured nature, SQuAD exhibits a high baseline retrieval accuracy, making it a useful testbed to assess whether metadata enrichment can deliver measurable improvements even under near-optimal conditions.

As shown in Table \ref{tab:retrieval-improvements} and Figure~\ref{fig:SQuAD_accuracy}, the baseline retrieval accuracy without metadata is already high at 93.30\%. Nevertheless, enriching the dataset with MetaGen metadata yields a further improvement to 93.68\%. While this 0.38 percentage point gain may appear small, it translates to 39 additional correct retrievals on the 10\,570-question SQuAD development set—demonstrating that even marginal gains in highly saturated regimes can be practically meaningful.

This result reinforces the value of metadata enrichment as a universally applicable technique: it remains beneficial not only in metadata-scarce or low-performing domains like biomedical QA but also in high-performing datasets where further improvements are difficult to achieve.

Overall, our retrieval experiments across both generic and structured datasets affirm the efficacy and generalizability of the proposed metadata enrichment pipeline. 

\begin{table}[h]
\caption{Retrieval Accuracy Before and After MetaGen}
\label{tab:baseline-vs-best-enriched}
\centering
\begin{tabular}{@{}lcc@{}}
\toprule
\textbf{Dataset} & \textbf{Without metadata (\%)} & \textbf{After MetaGen enrichment (\%)} \\
\midrule
PubMedQA & 74.9 & \textbf{82.1} \\
Natural Questions (NQ) & 49.99 & \textbf{60.48} \\
SQuAD & 93.3 & \textbf{93.68} \\
\bottomrule
\end{tabular}
\end{table}

\subsection{Retrieval-Augmented Generation (RAG) Evaluation}

To evaluate the downstream benefits of our proposed metadata enrichment strategy, we conducted extensive experiments integrating enriched retrieval with a standard RAG framework on the PubMedQA dataset. We have used the Llama3.1-70 B-instruct model for answer generation.\cite{grattafiori2024llama}

\begin{figure}[h]
\centering
\includegraphics[width=0.8\linewidth]{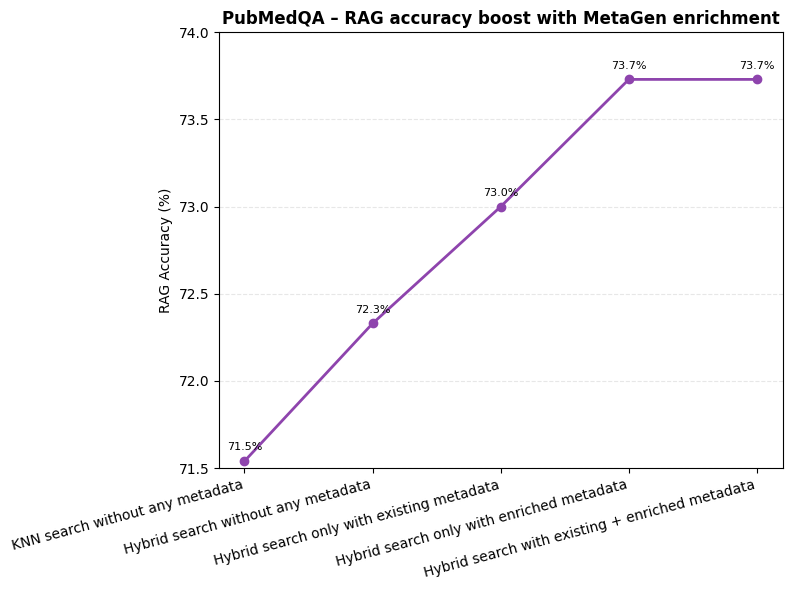}
\caption{Impact of MetaGen Metadata Enrichment on RAG Accuracy (PubMedQA Dataset)}
\label{fig:MBRA}
\end{figure}

Figure \ref{fig:MBRA} systematically illustrates the incremental improvements in RAG performance resulting from our enriched retrieval approach. Using the baseline retrieval pipeline equipped only with existing metadata available in PubMedQA, we observed a short-answer accuracy of approximately 71\%. Incorporating passages retrieved using our metadata-enriched indexing substantially elevated this accuracy to approximately 74\%, representing an absolute improvement of around 3 percentage points, corresponding to a relative performance gain of approximately 5\%. Crucially, these improvements were achieved exclusively through enhanced retrieval—without any modifications to the underlying model or prompt optimization—demonstrating the clear and direct benefit of providing semantically richer and more contextually relevant retrieval results to the generative model.

The primary mechanism underlying this performance improvement involves enriching retrieval contexts with medically relevant semantic details, including specialized synonyms (e.g., "MI" for "myocardial infarction"), domain-specific acronyms, and precise topical phrases frequently used by medical practitioners. By explicitly embedding this medical domain knowledge into the retrieval stage, our enrichment pipeline significantly enhances the retriever’s capability to identify and present the most pertinent documents, enabling the generative model to produce more accurate and contextually grounded answers.

Table \ref{pubmedqa-nonfinetuned-table} and Table \ref{pubmedqa-finetuned-table} benchmarks our MetaGen BlendedRAG results against recent state-of-the-art approaches evaluated on the PubMedQA dataset. Our approach first demonstrates superior performance compared to several competitive approaches of retrievalrs. Furthermore, even though our method is non-fine-tuned, we showed much better performance across several domain-specific fine-tuning approaches. Specifically, our metadata-driven pipeline achieves an accuracy of 77.9\%, substantially surpassing widely recognized methods such as AlzheimerRAG (74\%) \cite{lahiri2024alzheimerrag}, RAFT with LLaMA2-7B (73.3\%) \cite{zhang2024raft}, and GPT-3.5 integrated RAG systems (71.6\%). Our MetaGen pipeline notably achieves best performance without relying on extensive fine-tuning. Instead, the observed improvements stem directly from a sophisticated metadata enrichment strategy that significantly enhances retrieval effectiveness, emphasizing the critical role metadata quality plays in RAG systems' downstream generative accuracy.

These empirical results compellingly demonstrate that a systematic and targeted metadata enrichment strategy not only significantly boosts retrieval precision but also directly translates into substantial generative accuracy gains in RAG pipelines. 

\begin{table}[h]
\caption{RAG Accuracy on PubMedQA Across Non-Fine-Tuned Models}
\label{pubmedqa-nonfinetuned-table}
\centering
\begin{tabular}{@{}l l c@{}}
\toprule
\textbf{Paper (Approach)} & \textbf{Description} & \textbf{Accuracy (\%)} \\
\midrule
\textbf{MetaGen BlendedRAG} & \textbf{Leverages metadata augmentation for retrieval} & \textbf{77.9} \\
GPT-3.5 + RAG\cite{zhang2024raft} & RAG pipeline with GPT-3.5 model & 71.6 \\
LLaMA2-7B + RAG\cite{zhang2024raft} & RAG pipeline with LLaMA2-7B model & 58.8 \\
\bottomrule
\end{tabular}
\end{table}

\begin{table}[h]
\caption{RAG Accuracy on PubMedQA Across Fine-Tuned Models}
\label{pubmedqa-finetuned-table}
\centering
\begin{tabular}{@{}l p{7cm} c@{}}
\toprule
\textbf{Paper (Approach)} & \textbf{Description} & \textbf{Accuracy (\%)} \\
\midrule
RankRAG\cite{yu2024rankrag} & Fine-tuned RAG pipeline for domain & 79.80 \\
\textbf{MetaGen BlendedRAG} & \textbf{Leverages metadata augmentation for retrieval (Non Fine-Tuned)} & \textbf{77.9} \\
AlzheimerRAG\cite{lahiri2024alzheimerrag} & Multimodal integration (text + visuals) & 74.0 \\
RAFT (LLaMA2-7B)\cite{zhang2024raft} & Retrieval-augmented fine-tuning & 73.3 \\
DSF + RAG & Domain-specific fine-tuning for RAG & 71.6 \\
MEDRAG + GPT-4 & Corpus-optimized medical retrieval & 70.6 \\
\bottomrule
\end{tabular}
\end{table}

\section{Conclusion \& Future Work}

In conclusion, the MetaGen Blended Retrieval-Augmented Generation (MBRAG) approach provides a broadly applicable methodology for improving performance across diverse, domain-specific RAG.
It offers a scalable and domain-adaptive solution to address the persistent challenge of low retrieval accuracy in metadata-deficient corpora. By introducing a conditional metadata enrichment framework, along with a hybrid retrieval strategy that fuses lexical (BM25) and semantic (dense vector) indexing, MBRAG enhances context selection quality in RAG systems—without requiring expensive model fine-tuning.

Empirical results validate the effectiveness of our approach across three benchmark datasets: PubMedQA, Natural Questions (NQ), and SQuAD. On the PubMedQA dataset, MBRAG achieves a top-1 retrieval accuracy of 82.1\%, a substantial improvement over the baseline of 74.9\%. This lift is driven by progressive metadata integration, with the most effective configuration involving a boosted hybrid search over both existing and enriched metadata. Similarly, in the NQ dataset, where baseline accuracy was much lower at 49.99\%, metadata enrichment raises retrieval performance to 60.48\%—a 10.5-point absolute gain. Even in the high-performing SQuAD dataset, where the baseline without metadata is 93.3\%, enriched metadata leads to a new peak of 93.68\%, equating to 39 additional correct retrievals out of the 10,570-question dev set.

These results confirm that the MetaGen enrichment pipeline consistently improves retrieval accuracy across both high- and low-baseline datasets. The gains are especially meaningful in domain-specific tasks, where high retrieval precision is critical for downstream answer generation. Moreover, since MBRAG operates entirely without model fine-tuning, it provides a cost-efficient and infrastructure-friendly alternative suitable for enterprise and real-world deployments.

The approach is not without limitations and shows incremental and diminishing returns for a pre-trained model on domain data. In the future, the method can be combined with fine-tuning to produce even more accurate models for domain-specific QA, and an automated metadata detector for important metadata for a specific domains can be constructed.

\bibliographystyle{unsrt}  
\bibliography{ref}

\newpage
\section{Appendix}
\subsection{Dataset information}
\subsubsection{PubMedQA Dataset}
\begin{figure}[H]
\centering
\includegraphics[width=1\linewidth]{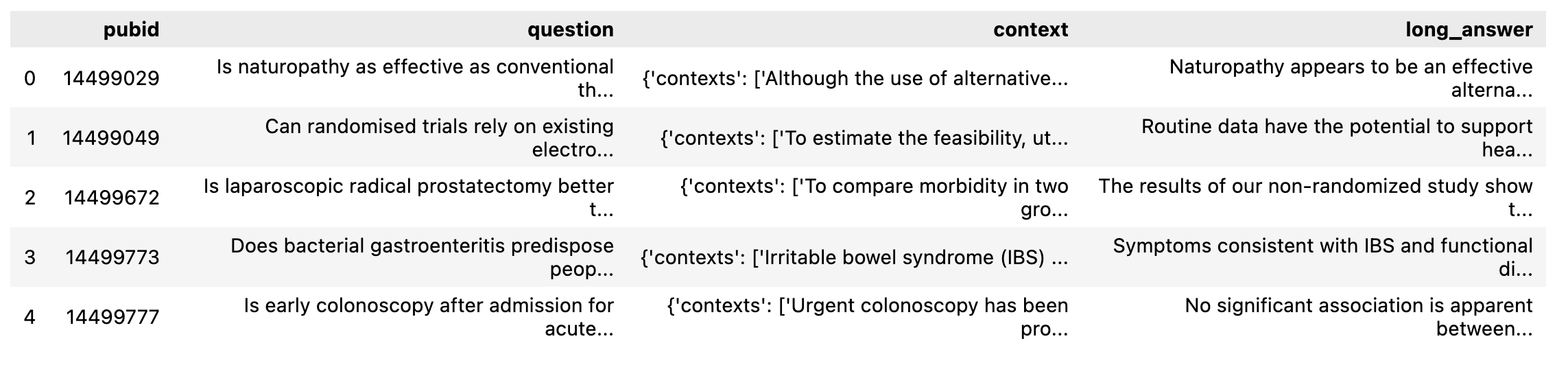}
\caption{PubMedQA sample data}
\label{fig:PubMedQAdataset}
\end{figure}

We downloaded the PubMedQA dataset from \href{https://huggingface.co/datasets/qiaojin/PubMedQA} {huggingface}. PubMedQA is a biomedical question answering dataset created to advance research in scientific and medical QA. The dataset is built from PubMed abstracts, where each instance consists of a research question (often derived from article titles), a context (the abstract without its conclusion), a long answer (the abstract’s conclusion), and a yes/no/maybe answer summarizing the conclusion as shown in Figure \ref{fig:PubMedQAdataset}. PubMedQA includes 1,000 expert-annotated and 61,249 unlabeled QA pairs, making it a valuable resource for developing and evaluating models that require reasoning over biomedical research texts.
\subsubsection{PubMedQA Corpus Details}
To support robust benchmarking and reproducibility of research, we are sharing two versions of the PubMedQA dataset for the community that was used for this study. 

\begin{itemize}
    \item \textbf{PubMedQA\_original\_corpus\_combined.json:} This file is constructed directly from the official PubMedQA dataset, preserving its original structure. Each record contains a biomedical question, a context passage (abstract), and the corresponding answer, providing a strong baseline for QA and retrieval tasks in the biomedical domain. We have constructed this file to test the efficacy of the retriever in finding the right document and passage across using the method.

    In total, we have 62,249 documents combined in the corpus. The retriever now has a much harder job of searching for the right document. This is a harder RAG task than when a context is passed, as it is done on other benchmarks like GPT-3.5 + RAG \cite{zhang2024raft} and LLaMA2-7B + RAG \cite{zhang2024raft}.

    \begin{figure}[H]
    \centering
    \includegraphics[width=1\linewidth]{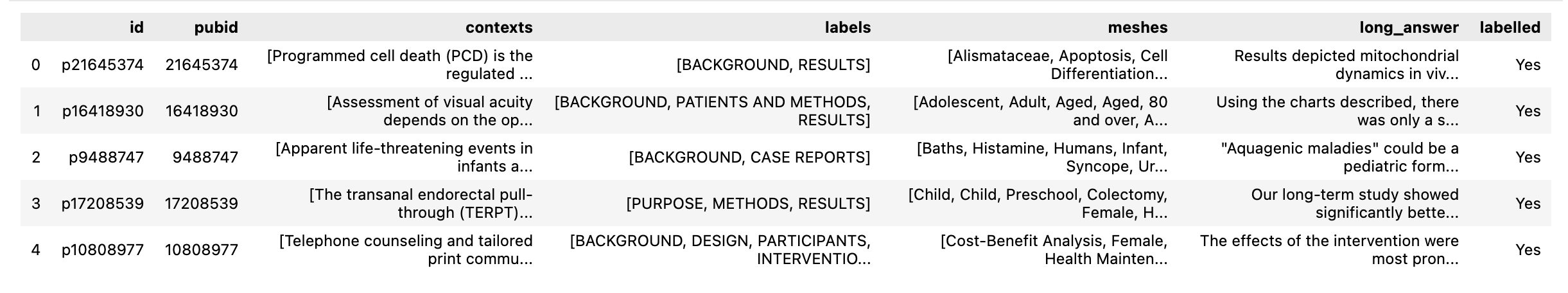}
    \caption{PubMedQA sample combined data}
    \label{fig:PubMedQAdataset Input}
    \end{figure}
    
    \item \textbf{PubMedQA\_corpus\_with\_metadata.json:} This is a metadata-enriched corpus created by our MetaGenBlendedRAG pipeline. In addition to the original fields, each entry is augmented with structured metadata, including keywords, topics, key phrases, synonyms, and acronyms. This enrichment is designed to improve document retrievability, enhance semantic search, and facilitate context-aware question answering.

\end{itemize}

\begin{figure}[H]
\centering
\includegraphics[width=1\linewidth]{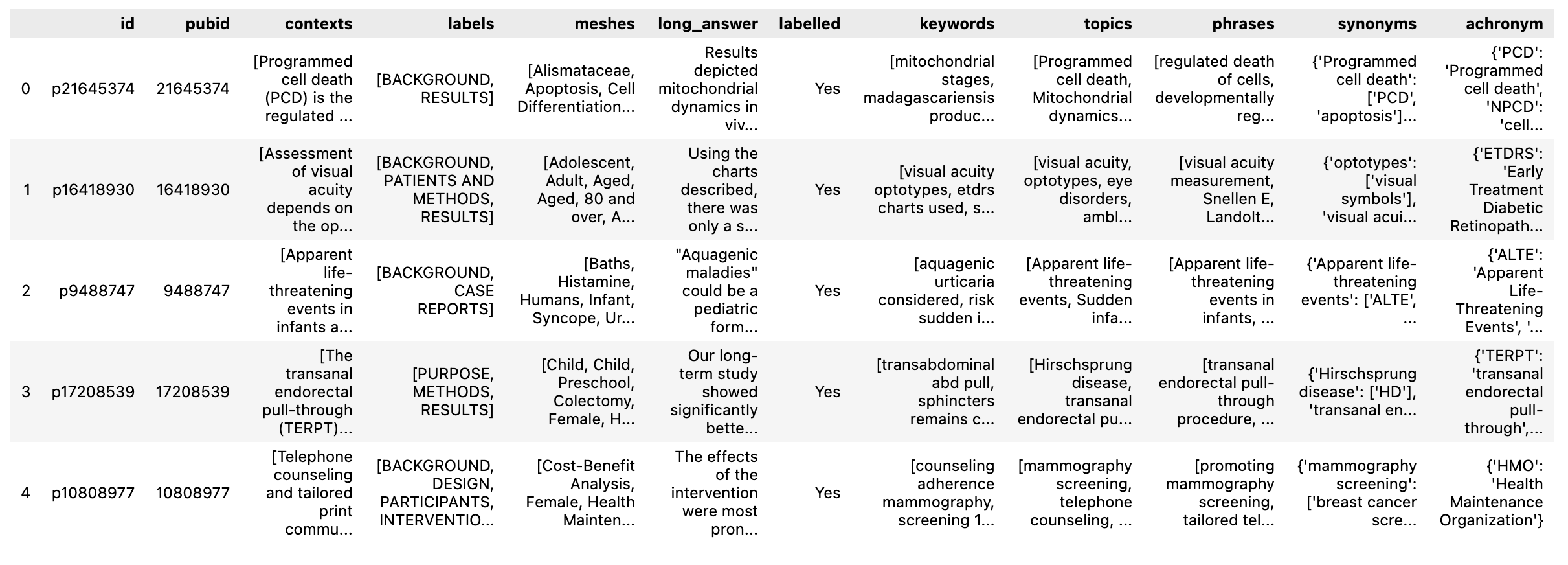}
\caption{PubMedQA-MetaGen: Example of metadata-enriched PubMedQA entry}
\label{fig:PubMedQAEnrichedCorpus}
\end{figure}

Both datasets are made publicly available via  \href{https://huggingface.co/datasets/Shivam6693/PubMedQA-MetaGenBlendedRAG}{HuggingFace} to benefit the broader research community and enable reproducible evaluation of both traditional and metadata-driven retrieval methods. The enriched corpus is illustrated in Figure \ref{fig:PubMedQAEnrichedCorpus}, showing the expanded metadata fields that accompany each biomedical entry.

\subsubsection{NQ Dataset}
\begin{figure}[H]
\centering
\includegraphics[width=1\linewidth]{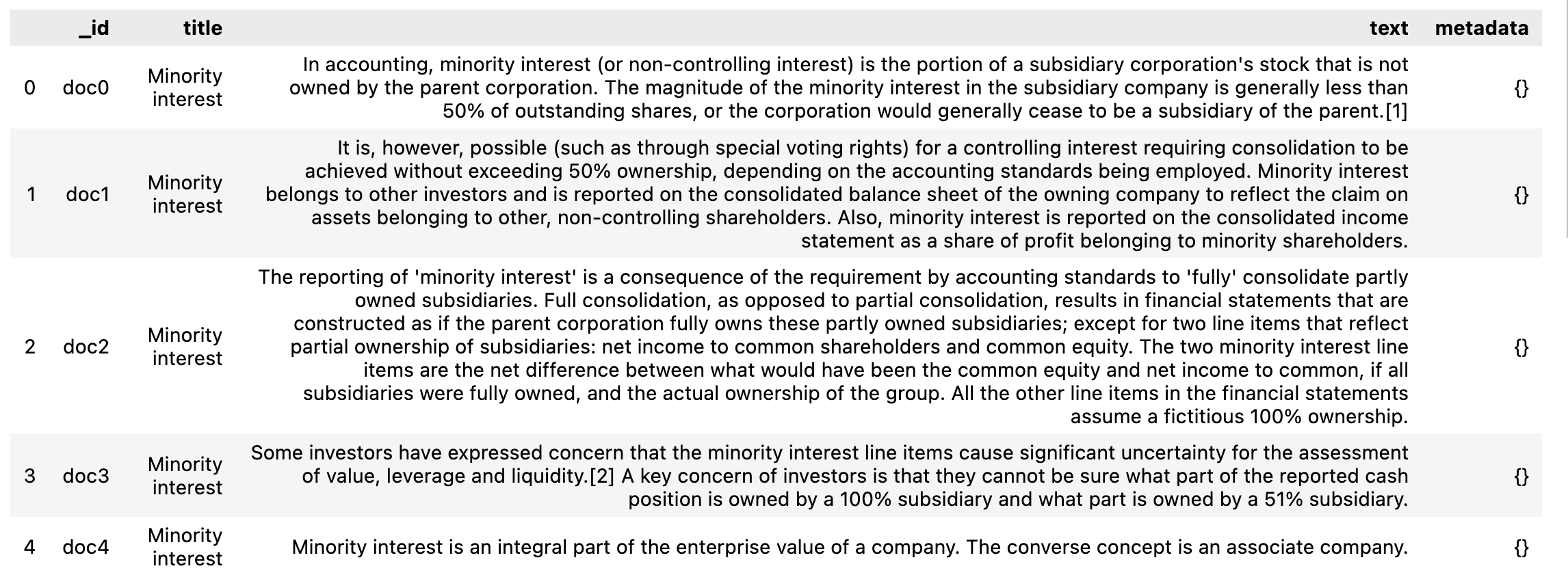}
\caption{NQ sample data}
\label{fig:NQ_dataset}
\end{figure}

We downloaded this data from \href{https://github.com/beir-cellar/beir}{GitHub Bier}. This dataset is created by Google AI, which is available for open source to help out open-domain question answering; the NQ corpus as shown in Figure \ref{fig:NQ_dataset} contains questions from real users, and it requires QA systems to read and comprehend an entire Wikipedia article that may or may not contain the answer to the question. 

\subsubsection{Squad Dataset}
\begin{figure}[H]
\centering
\includegraphics[width=1\linewidth]{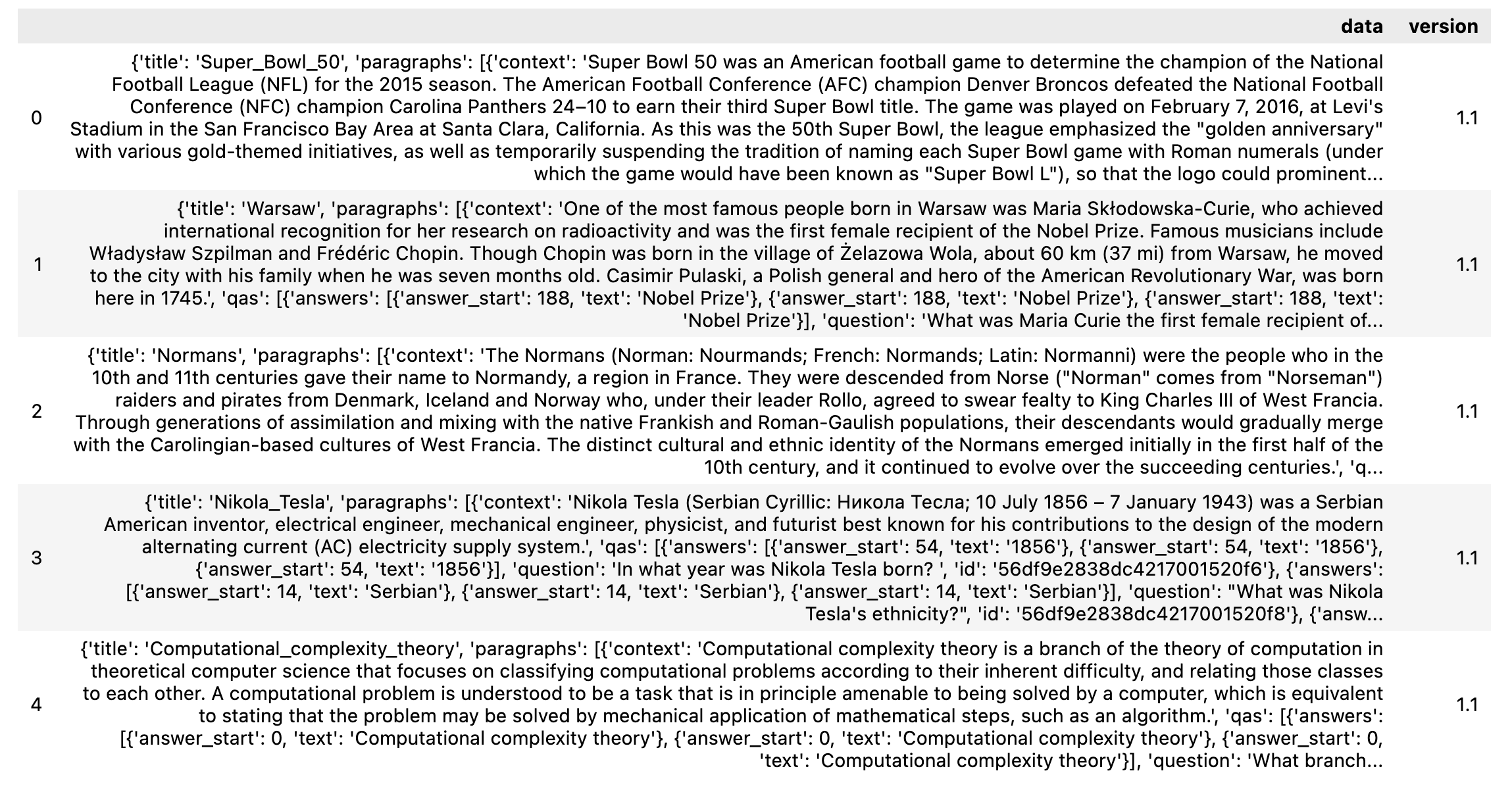}
\caption{Squad sample data}
\label{fig:squad_dataset}
\end{figure}

Stanford Question Answering Dataset(SQuAD) dataset is an open-source large-scale dataset. It is a collection of question-answer pairs derived from Wikipedia articles. There are two datasets available: squad 1.1 and Squad 2.0. We used the \href{https://rajpurkar.github.io/SQuAD-explorer/}{squad 1.1 } dataset. This data contains dev and train datasets. We used the dev dataset for our experiments.  This dataset as shown in Figure \ref{fig:squad_dataset} contains 2067 documents and 10570 question-answer pairs. 

\subsection{Detailed results}

\subsubsection{Retriever Results}

\begin{figure}[H]
\centering
\includegraphics[width=0.7\linewidth]{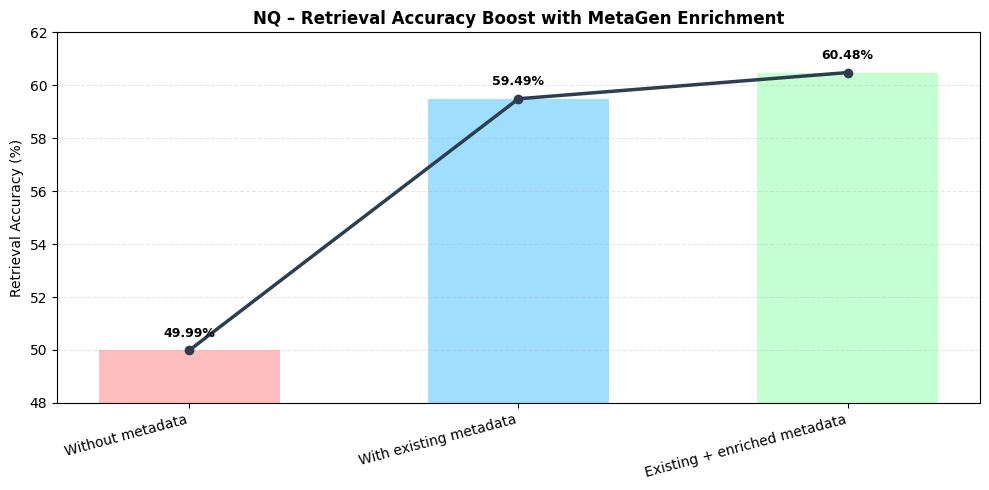}
\caption{Impact of MetaGen Metadata Enrichment on Top-5 Retrieval Accuracy (NQ Dataset)}
\label{fig:NQ_accuracy}
\end{figure}

The NQ evaluation as shown in Figure \ref{fig:NQ_accuracy} reveals that augmenting documents with the existing \textit{title} field alone lifts retrieval accuracy from \(\approx 50\%\) to \(\approx 59.5\%\) (a \(+9.5\)‑point, \(\sim 19\%\) relative gain).  
Adding MetaGen’s enriched KeyBERT tags delivers an additional boost, reaching \(\approx 60.5\%\) and illustrating the incremental—but tangible—benefit of automated metadata enrichment on top of conventional fields.

\begin{figure}[H]
\centering
\includegraphics[width=0.7\linewidth]{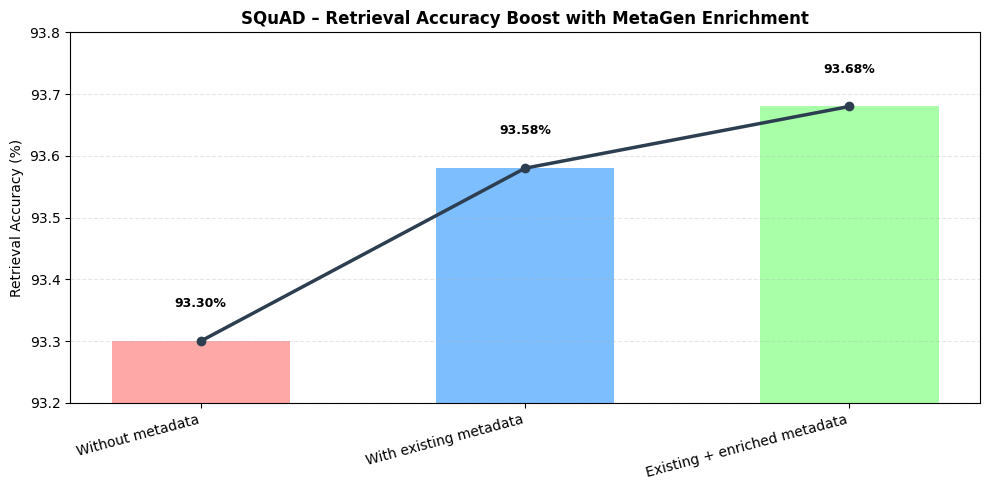}
\caption{Impact of MetaGen Metadata Enrichment on Top-5 Retrieval Accuracy (SQuAD Dataset)}
\label{fig:SQuAD_accuracy}
\end{figure}

For SQuAD, retrieval accuracy rises from \(\textbf{93.30\%}\) with no metadata to \(\textbf{93.58\%}\) when the existing \textit{title} field is indexed, and reaches \(\textbf{93.68\%}\) once MetaGen’s enriched topic tags are added as shown in Figure \ref{fig:SQuAD_accuracy}.  
Although the absolute gain is only \(+0.38\) percentage points, on the 10\,570‑question dev set this translates to roughly \(39\) additional passages retrieved correctly, underscoring how even marginal improvements matter at high‑baseline performance levels.

\subsubsection{RAG Results}
As illustrated in Table \ref{tab:rag-improvements}, the integration of MetaGen enrichment substantially improved the performance of the retrieval-augmented generation (RAG) framework. The inclusion of enriched metadata within the RAG pipeline yielded a notable absolute increase in accuracy relative to the baseline configuration that operated without metadata. This marked improvement underscores the effectiveness and generalizability of the proposed method in different datasets.

\begin{table}[H]
\caption{RAG accuracy improvements using MetaGen RAG across datasets}
\label{tab:rag-improvements}
\centering
\begin{tabular}{@{}p{1.8cm}p{7.2cm}c@{}}
\toprule
\textbf{Dataset} & \textbf{Search + Metadata Variant} & \textbf{RAG Accuracy (\%)} \\
\midrule
\multirow{8}{*}{\textbf{PubMedQA}} 
    & KNN search without any metadata & 71.5 \\
    & Hybrid search without any metadata & 72.33 \\
    & Hybrid search only with existing metadata & 73.00 \\
    & Hybrid search only with enriched metadata & 73.73 \\
    & Hybrid search with existing + enriched metadata & 73.73 \\
    & Hybrid (boosted) with enriched metadata & 73.54 \\
    & Hybrid (boosted) with existing metadata & 74.53 \\
    & Hybrid (boosted) with existing + enriched metadata & \textbf{77.9} \\
\midrule
\multirow{3}{*}{\textbf{NQ}} 
    & Without metadata & 24.77 \\
    & With existing metadata & 27.42 \\
    & Existing + enriched metadata & \textbf{26.71} \\
\midrule
\multirow{3}{*}{\textbf{SQuAD}} 
    & With existing metadata & 57 \\
    & Existing + enriched metadata & \textbf{58.50} \\
\bottomrule
\end{tabular}
\end{table}

\subsection{Github}
The source code associated with this paper is publicly available on GitHub and can be accessed at the following repository: \href{https://github.com/ibm-self-serve-assets/MetaGen-Blended-RAG}{MetaGen-Blended-RAG}.

\section{Acknowledgment}
The authors would like to thank the IBM Ecosystem \& Client Engineering team and leadership for their support in making this study possible. This work was inspired by real challenges faced by clients in scaling RAG. 

\newpage

\end{document}